\documentclass[10pt,twocolumn,letterpaper]{article}

\usepackage{caption}

\usepackage{cvpr}
\usepackage{times}
\usepackage{epsfig}
\usepackage{graphicx}
\usepackage{amsmath}
\usepackage{amssymb}
\usepackage{subfigure}

\usepackage[pagebackref=true,breaklinks=true,letterpaper=true,colorlinks,bookmarks=false]{hyperref}

\cvprfinalcopy 

\def\cvprPaperID{****} 

\newcommand\imgwidth{0.11}

\ifcvprfinal\fi
\begin{document}

\title{Understanding Beauty via Deep Facial Features}

\author{Xudong Liu, Tao Li, Hao Peng, Iris Chuoying Ouyang, Taehwan Kim, and Ruizhe Wang\\
ObEN, Inc\\
{\tt \{xudong,tao,hpeng,iris,taehwan,ruizhe\}@oben.com}\\
}

\twocolumn[{%
\renewcommand\twocolumn[1][]{#1}%
\maketitle
\begin{center}
\includegraphics[width=\imgwidth\linewidth]{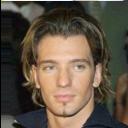}
\includegraphics[width=\imgwidth\linewidth]{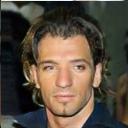}
~
\includegraphics[width=\imgwidth\linewidth]{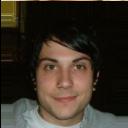}
\includegraphics[width=\imgwidth\linewidth]{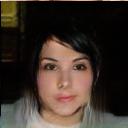}
~
\includegraphics[width=\imgwidth\linewidth]{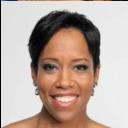}
\includegraphics[width=\imgwidth\linewidth]{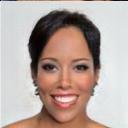}
~
\includegraphics[width=\imgwidth\linewidth]{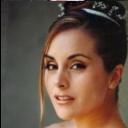}
\includegraphics[width=\imgwidth\linewidth]{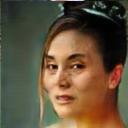}
\\
\includegraphics[width=\imgwidth\linewidth]{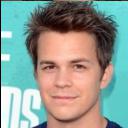}
\includegraphics[width=\imgwidth\linewidth]{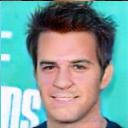}
~
\includegraphics[width=\imgwidth\linewidth]{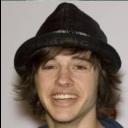}
\includegraphics[width=\imgwidth\linewidth]{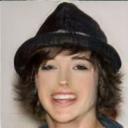}
~
\includegraphics[width=\imgwidth\linewidth]{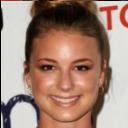}
\includegraphics[width=\imgwidth\linewidth]{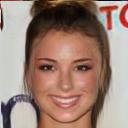}
~
\includegraphics[width=\imgwidth\linewidth]{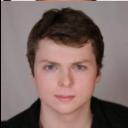}
\includegraphics[width=\imgwidth\linewidth]{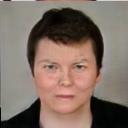}
\\
\includegraphics[width=\imgwidth\linewidth]{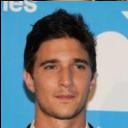}
\includegraphics[width=\imgwidth\linewidth]{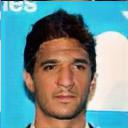}
~
\includegraphics[width=\imgwidth\linewidth]{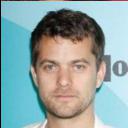}
\includegraphics[width=\imgwidth\linewidth]{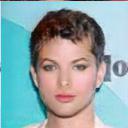}
~
\includegraphics[width=\imgwidth\linewidth]{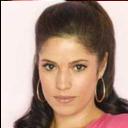}
\includegraphics[width=\imgwidth\linewidth]{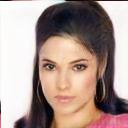}
~
\includegraphics[width=\imgwidth\linewidth]{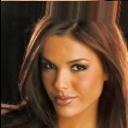}
\includegraphics[width=\imgwidth\linewidth]{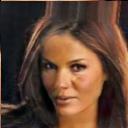}
\\
\includegraphics[width=\imgwidth\linewidth]{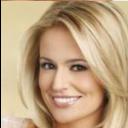}
\includegraphics[width=\imgwidth\linewidth]{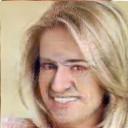}
~
\includegraphics[width=\imgwidth\linewidth]{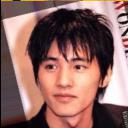}
\includegraphics[width=\imgwidth\linewidth]{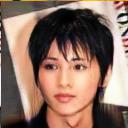}
~
\includegraphics[width=\imgwidth\linewidth]{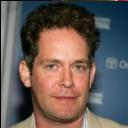}
\includegraphics[width=\imgwidth\linewidth]{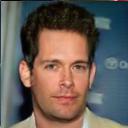}
~
\includegraphics[width=\imgwidth\linewidth]{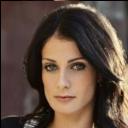}
\includegraphics[width=\imgwidth\linewidth]{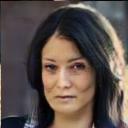}
\captionof{figure}{\textbf{\textit{In each image pair, which one (Left or Right) is more attractive?}} We propose a method and a novel perspective of beauty understanding via deep facial features, which allows us to analyze which facial attributes contribute positively or negatively to beauty perception. To validate our result, we manipulate the facial attributes and synthesize new images. In each case, left corresponds to the original image, and right represents the synthesized one. The sample modified facial attributes from left to right are small nose to big nose, male to female, no-makeup to makeup, and young to aged. To see our discovery to the first question, please read remaining of the paper.}
\label{fig:GAN}
\end{center}

}]

\vspace{10pt}
\begin{abstract}
\vspace{-10pt}
The concept of beauty has been debated by philosophers and psychologists for centuries, but most definitions are subjective and metaphysical, and deficit in accuracy, generality, and scalability. In this paper, we present a novel study on mining beauty semantics of facial attributes based on big data, with an attempt to objectively construct descriptions of beauty in a quantitative manner.
We first deploy a deep Convolutional Neural Network (CNN) to extract facial attributes, and then investigate correlations between these features and attractiveness on two large-scale datasets labelled with beauty scores. Not only do we discover the secrets of beauty verified by statistical significance tests, our findings also align perfectly with existing psychological studies that, e.g., small nose, high cheekbones, and femininity contribute to attractiveness. We further leverage these high-level representations to original images by a generative adversarial network (GAN). Beauty enhancements after synthesis are visually compelling and statistically convincing verified by a user survey of 10,000 data points.
\end{abstract}

\section{Introduction}\label{sec:introduction}
\begin{figure*}[t]
\begin{center}
\includegraphics[width=1.0\textwidth]{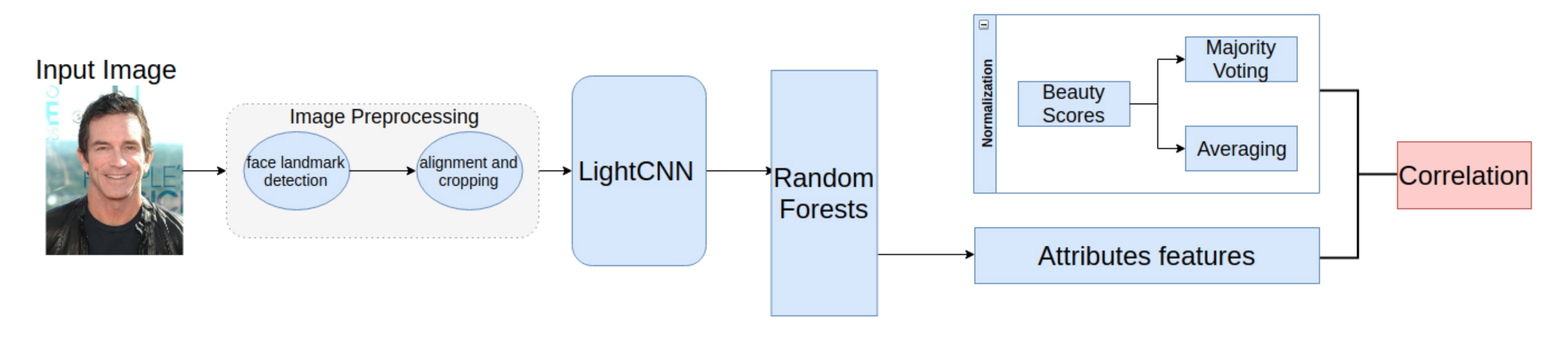}
\end{center}
\caption{An overview of our approach.}
\label{fig:overview}
\end{figure*}

Facial attractiveness has profound effects on multiple aspects of human social activities, from inter-sexual and intra-sexual selections to hiring decisions and social exchanges \cite{little2011facial}.
For example, facially attractive people enjoy higher chances of getting dates \cite{riggio1984role} and their partners are more likely to gain satisfaction compared to dating with less attractive ones \cite{berscheid1971physical}.
Overwhelmed by social fascination with beauty, less facially attractive people might suffer from social isolation, depression, and even psychological disorders \cite{bull2012social,rankin1998quality,phillips1993body,macgregor1989social,bradbury1994psychology}.
{\em Cash et al.} \cite{cash1985aye} found that attractive people are in better positions when finding jobs. Attractiveness of a suspect can even impact the judge's decision \cite{sigall1975beautiful}.

Over centuries, studies of facial beauty have attracted consistent interest among psychologists, philosophers, and artists, the majority of whom focus on human perception.
What is beauty? Psychologists response to this question by investigating various factors, ranging from symmetry \cite{thornhill1994human,mealey1999symmetry,rhodes1998facial,scheib1999facial} and averageness \cite{galton1878composite,langlois1990attractive,rhodes1996averageness} to personality \cite{little2006good} and sexual dimorphism \cite{rhodes2003does,penton2004high}.

Although have been studied extensively in the psychology community, studies of beauty are relatively new to the computing world. With the popularity of digital cameras as well as social media, images are increasingly pervasive in almost all aspects of social life and many computational beauty enhancement methods have been proposed recently \cite{zhang2016new,leyvand2008data,arakawa2005system,melacci2010template,scherbaum2011computer,liao2012enhancing,zhang2017facial,liu2016advances,zhang2016beauty}, most of which rely on previous psychological findings. Their main idea behind is to analyze low-level geometric facial features (e.g., shape ratio, symmetry, texture) and then apply machine learning algorithms, such as support vector machines (SVMs) \cite{kagian2007humanlike,chen2010novel,mao2009automatic} and k-nearest neighbors (k-NN) \cite{aarabi2001automatic} to perform image classifications or beauty predictions \cite{fan2012prediction,gray2010predicting,schmid2008computation,nguyen2013towards,altwaijry2013relative,wang2014attractive}.
Features such as local binary patterns (LBP) \cite{ojala2002multiresolution} and Gabor \cite{whitehill2008personalized,chen2010novel} are extracted to train auto-raters in supervised manner, where beauty scores are collected and labelled manually.

Instead of using low-level facial geometric features based on psychological findings, we propose a novel study of correlations between facial attractiveness and facial attributes (e.g., shape of eyebrows, nose size, hair color), inspired by {\em Leyvand et al.} \cite{leyvand2008data} who suggested that high-level facial features play critical roles in beauty estimation.
Our study is driven by the explosion of big data as well as promising performance of deep learning models. As illustrated in Figure \ref{fig:overview}, we first deploy a deep Convolutional Neural Network (CNN) for facial attribute estimation. Correlations between high-level features and beauty outcomes are then studied in two large-scale datasets of labelled real-world images \cite{zhang2016new,bainbridge2013intrinsic}.
Facial attributes showing statistically significant correlations with beauty outcomes are thereby selected. We further correlate our results with psychological findings, and discuss their similarities and differences. In the end, we integrate above attributes with a generative adversarial network (GAN) to generate beautified images, which demonstrate perceptually appealing outcomes and validate the correctness our study as well as previous psychological works.


Major contributions in this paper include:
\begin{itemize}
    \item We extract facial attributes using deep CNNs trained in two large-scale real-world datasets labelled with beauty scores.
    \item We are the first to objectively analyze correlations between beauty and facial attributes with a quantitative approach and select statistically significant attributes of attractiveness.
    \item We validate existing psychological studies of beauty and discover new patterns.
    \item We integrate these facial features with a GAN to generate beautified images and then conduct a user survey of $10,000$ data points to verify the results.
\end{itemize}

The rest of the paper is organized as follows: Section \ref{sec:related} investigates previous works in facial attractiveness understanding. Section \ref{sec:method} describes our novel approach. Experiments are detailed in Section \ref{sec:experiment}. Section \ref{sec:analysis} presents further analysis. We concludes the paper in Section \ref{sec:conclusion}.

\section{Related Work}\label{sec:related}

In this section, we investigate existing studies of beauty from both psychological and computational perspectives.

\subsection{Psychological Studies}\label{sec:psycho}

What is beauty? This question has been debated by philosophers and psychologists for centuries. The well-known saying {\em beauty is in the eye of the beholder} indicates that the perception of beauty is subjective and non-deterministic as it stems from various cultural and social environments. However, cross-cultural agreements on facial attractiveness have been found in many studies \cite{cunningham1995their,langlois2000maxims,eisenthal2006facial}. In other words, people from diverse backgrounds around the globe share certain common criteria for beauty.

Many factors have been investigated by psychologists, including symmetry, averageness, and sexual dimorphism. {\em Rhodes et al.} \cite{rhodes1998facial} and {\em Perrett et al.} \cite{perrett1999symmetry} reported that symmetry has a positive influence on attractiveness. {\em Galton et al.} \cite{galton1878composite} noted that multiple faces blended together are more attractive than constituent faces, indicating that averaging face is another positive factor.
Several studies \cite{little2002role,perrett1998effects,rhodes2000sex,little2001self,little2002partnership} showed that people prefer feminine-looking faces regardless of actual genders of the faces.
{\em Hassin et al.} \cite{hassin2000facing} found that smiling faces are more attractive, which aligns with our intuitions.

\subsection{Computational Analysis}\label{sec:compute}

Secrets of beauty have been discussed in the psychology community for centuries; however, computer scientists didn't involve in this field until recent years. The fact that facial attractiveness plays such a pivotal role in the society as well as recent advances in computer vision motivate more and more researchers to involve, leading to a recent outburst of related products, such as mobile applications.

\begin{figure*}[t]
\begin{center}
\includegraphics[width=0.7\textwidth,height=0.2\textheight]{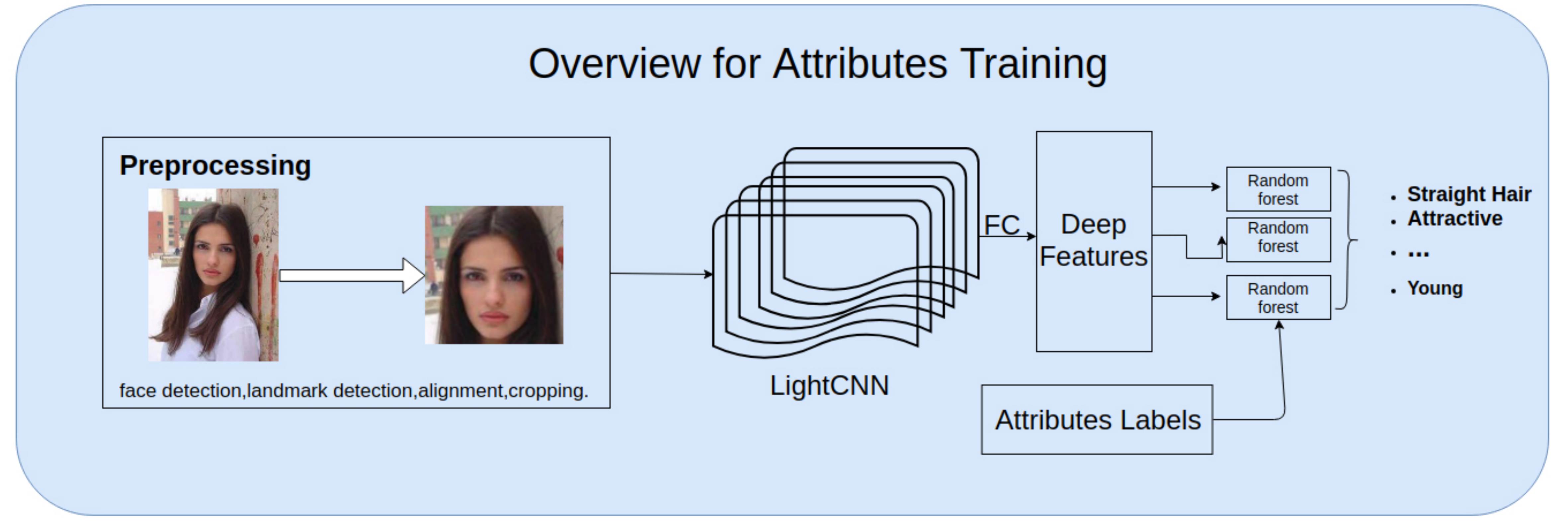}
\end{center}
\caption{An overview of attributes training}
\label{fig:training}
\end{figure*}

Numbers of researchers have demonstrated their contribution on how to beautify still images and predict facial beauty. {\em Chen et al.} \cite{zhang2016new} proposed a hypothesis on facial beauty perception. They found out that weighted averages of two geometric features are better and adopt their hypotheses on beautification model using SVR and have achieved the state-of-the-art geometric feature based face beautification. {\em Liu et al.} \cite{liu2017landmark} presented a purely landmark based, data-driven method to compute three kinds of geometric facial features for a 2.5D hybrid attractiveness computational model. A facial skin beautification framework to remove facial spots based on layer dictionary learning and sparse representation proposed by {\em Lu et al.} \cite{lu2016facial}. {\em Leyvan et al.} \cite{leyvand2008data} focused on enhancing the attractiveness of human faces in frontal view.
They presented face warping towards the beauty-weighted average of the $k$ closer samples in the face space. They also proposed that a small local adjustment can lead to an appreciable impact on the facial attractiveness (partly enhance). These findings inspired us to figure out which parts are mostly related to attractiveness, with an attempt to decorate specific small pieces (e.g., eyes) instead of the entire face for beautification.
{\em Chen et al.} \cite{chen2016data} also addressed that high-level features are beneficial to beauty prediction, which further drives us to figure out which specific attributes affect the beauty. Such high-level representations can further be applied for beauty enhancements, which will be shown in Section \ref{sec:change}.

It is worth mentioning that we propose the following extensions of \cite{liu2018facial}: 1) For deep facial features extraction, instead of training GoogLeNet \cite{szegedy2015going} in an end-to-end fashion, we directly employ an off-the-shelf facial feature extractor LightCNN \cite{wu2018light}, trained on millions of images for the face recognition task, and then do a random forests training for attributes prediction. This greatly reduces the training time and generalizes our framework to potentially work with any other tasks; 2) We perform statistical analysis on the estimated Pearson's Correlation Coefficient to demonstrate that the mined facial beauty semantics are statistically significant; 3) We adopt a state-of-the-art multiple-domain image translation framework StarGAN \cite{choi2017stargan} to manipulate facial attributes and perform extensive user study to evaluate beauty differences. This helps us to further practically validate the discovered correlation between beauty and semantic facial attributes.

\section{Method}\label{sec:method}

Figure \ref{fig:overview} illustrates an overview of our approach: data preprocessing, attributes training, correlation analysis, and attribute translation. These procedures are detailed in this section.

\begin{table*}[t]
\begin{center}
\begin{tabular}{ c|c|c|c|c|c|c}
\hline
$\textbf{Database}$  & $\textbf{Size}$   &$\textbf{Ethnicity}$  &$\textbf{Gender}$ & $\textbf{Score Scale}$   &$\textbf{Rating Number}$  &$\textbf{Normalization}$ \\
\hline
\hline
$\textrm{Beauty 799 \cite{zhang2016new}}$ &$\textrm{799}$ &$\textrm{Diverse}$ &$\textrm{Only Female}$  &$\textrm{3}$     &25   &$\textrm{Voting}$         \\
\hline
$\textrm{The 10k US \cite{bainbridge2013intrinsic}}$ &$\textrm{2222}$ &$\textrm{Caucasian}$ &$\textrm{Female and Male}$ &$\textrm{5}$     &12   &$\textrm{Averaging}$         \\
\hline
\end{tabular}
\end{center}
\caption{Beauty Dataset Description}
\label{tab:dataset_desc}
\end{table*}

\subsection{Data Preprocessing}\label{sec:preprocessing}

\begin{figure*}[t]
\subfigure[Image preprocessing]{\includegraphics[width=0.55\textwidth,height=0.15\textheight]{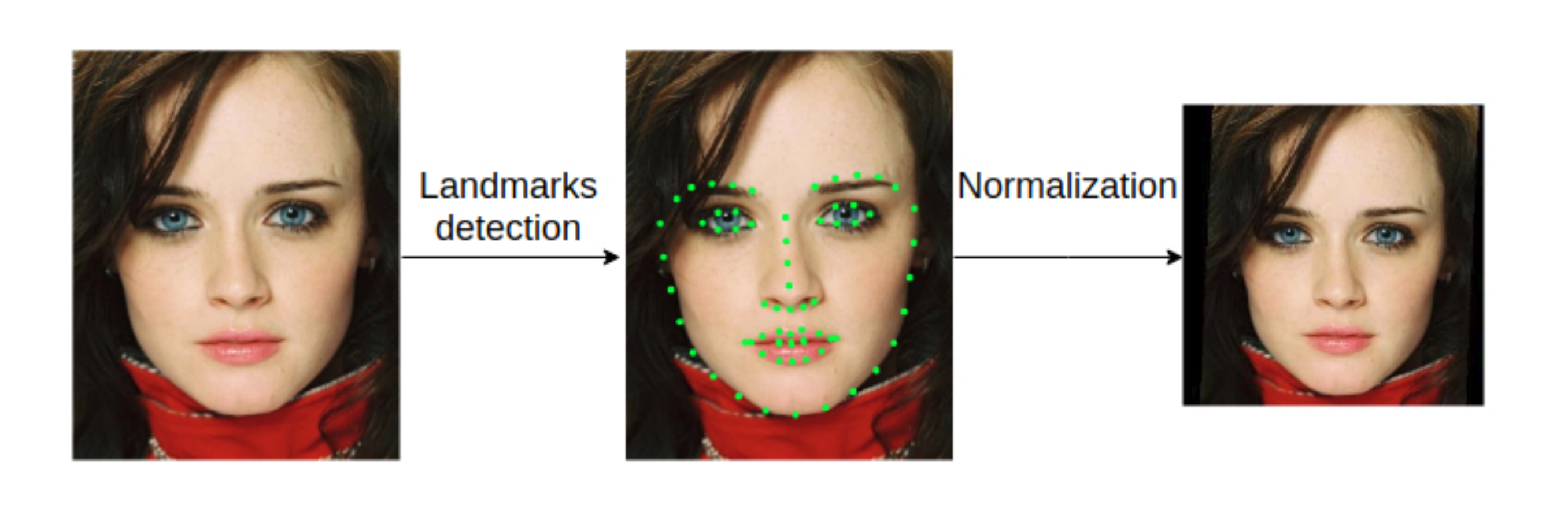}}
~
\subfigure[Facial attribute prediction]{\includegraphics[width=0.4\textwidth,height=0.2\textheight]{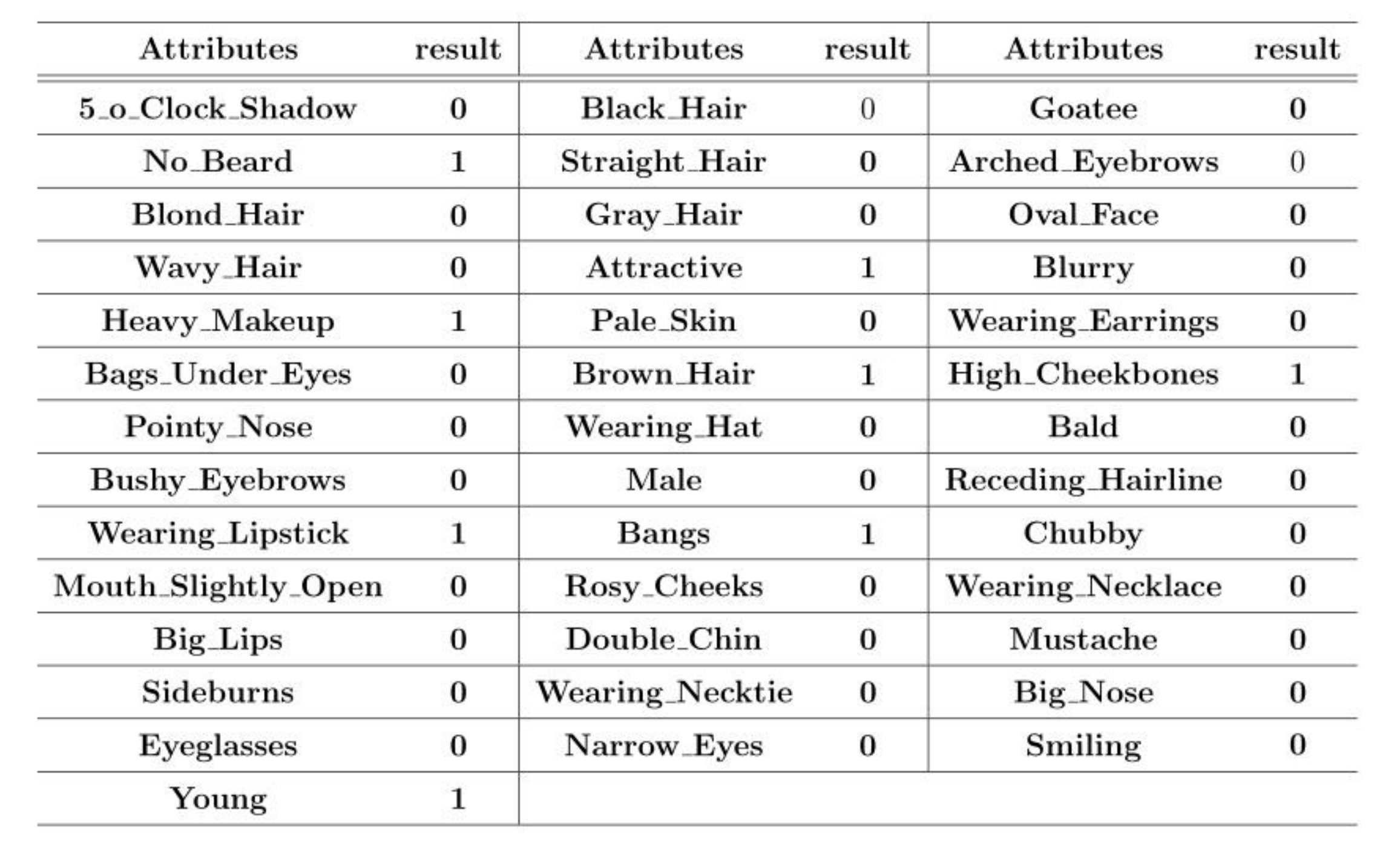}}
\caption{An example of image preprocessing and the corresponding attribute prediction results}
\label{fig:example}
\end{figure*}

Preprocessing is necessary for removing unwanted variation from the training data. There are four steps in our image normalization pipeline: face detection, landmarks detection, alignment, and cropping. OpenFace \cite{zadeh2017convolutional} is used for face and landmark detection. After detection, $68$ landmarks are provided as shown in Figure \ref{fig:example}. Given landmarks, face images are aligned and cropped with the size of $256\times256$.

In addition to image preprocessing, beauty scores also need normalization because they are inconsistent when multiple people rate per image and we adopt majority voting and averaging methods to generate scores from \cite{zhang2016new,bainbridge2013intrinsic}, respectively.

\subsection{Attributes Training}\label{sec:training}


In this paper, we employ LightCNN \cite{wu2018light} as the feature extractor. This network achieved state-of-the-art face recognition results on several face benchmarks, indicating the representations learned from it are convincing for feature extraction.


The overview of the training process is shown in Figure \ref{fig:training}. First, images are fed into LightCNN and features are extracted from Fully Connected (FC) layer. Then 40 random forest classifiers are trained for attributes estimation and finally output the attribute results.

\subsection{Correlation Analysis}\label{sec:corr_analysis}

After getting normalized beauty scores and the $40$ facial attributes, we perform an investigation for the secrets of beauty - correlation between facial attributes and beauty outcomes.

\subsubsection{Pearson's Correlation Coefficient}\label{sec:pearson}

Pearson's correlation coefficient (PCC) \cite{tutorials2014pearson} is used to measure linear relationship between two samples. It is calculated by
\begin{equation}\label{eqn:r}
r = \frac{\sum_{i=1}^n (x_i - \bar{x}) (y_i - \bar{y})}{\sqrt{\sum_{i=1}^n (x_i - \bar{x})^2} \sqrt{\sum_{i=1}^n (y_i - \bar{y})^2}}
\end{equation}
where $n$ is the sample size, $x_i$ and $y_i$ are sample points, and
\begin{align}
\bar{x} = \frac{1}{n} \sum_{i=1}^n x_i, ~ &\bar{y} = \frac{1}{n} \sum_{i=1}^n y_i ~ .
\end{align}
$r$ ranges from ${-1}$ to $1$, where the strongest positive linear correlation is represented by $1$, while $0$ indicates no correlation, and ${-1}$ indicates the strongest negative linear correlation.

\subsubsection{Testing Significance of Correlation Coefficients}\label{sec:significance_corr}

Although the Pearson's correlation coefficient provides the strength and direction of linear relationship between two samples, we need to confirm that the relationship is strong enough to be useful. Therefore, we apply statistical tests to investigate significance of correlation between beauty outcomes and facials attributes, where hypothesis tests are formed as
\begin{align*}
    & H_0: r = 0 \\
    & H_1: r \ne 0
\end{align*}
where $H_0$ is null hypothesis, $H_1$ is alternate hypothesis, and the significance level $\alpha$ is set to be $0.05$. If the calculated $p$-value is less than $\alpha$, we conclude that the correlation is significant; otherwise, we accept $H_0$.

\subsubsection{Testing Differences between Means}\label{sec:welch}

{\em Hunter} \cite{hunter1997needed} reported that the empirical average error rate across psychological studies is $60\%$, which is much higher than the $5\%$ error rate of significant tests that psychologists think to be. Thus, we are very skeptical and careful about reporting any results. Besides testing the significance of correlation coefficients, we futher test the significance of differences between means using different methods.

For any given facial attribute $i$, we split images into two groups by attribute $i$. The null and alternative hypotheses are, respectively,
\begin{align*}
    & H_0: \mu_{i0} = \mu_{i1} \\
    & H_1: \mu_{i0} \ne \mu_{i1}
\end{align*}
where $\mu_{i0}$ denotes the average beauty score of the group without attribute $i$ and $\mu_{i1}$ denotes the mean score for the other group.

Independent two-sample $t$-tests \cite{kirk2007experimental} are widely used to compare whether the average difference between two groups is statistically significant or instead due to random effects. The $t$ statistic for equal sample sizes and equal variances is defined by
\begin{equation}
    t = \frac{\bar{X}_0 - \bar{X}_1}{\sqrt{\frac{s_{X_0}^2 + s_{X_1}^2}{2}}}
\end{equation}
where $s_{X_0}^2$ and $s_{X_1}^2$ are unbiased estimators of the variances of the two samples. However, the equivalence of sample sizes and variances are not guaranteed in our case. We futher introduce Welch's $t$-test \cite{welch1947generalization} to estimate variances separately. The $t$ statistic for Welch's t-test is calculated by
\begin{equation}
    t = \frac{\bar{X}_0 - \bar{X}_1}{\sqrt{ \frac{s_0^2}{n_0} + \frac{s_1^2}{n_1} }}
\end{equation}
where $s_0$ and $s_1$ are unbiased estimators of variances of each group. We set the significance level to $0.05$ and test in single-tailed manner. If the $p$-value is less than $0.05$, we reject $H_0$ and conclude a significant corelation between attribute $i$ and beauty score; otherwise, we accept $H_0$ which means that average beauty scores of the two groups have no significant difference.

\subsection{Attribute Translation}\label{sec:attr_translation}

To quantitative evaluate beauty differences with or without certain attributes, a generative adversarial network (GAN) \cite{goodfellow2014generative} is deployed to transfer facial attributes. GAN is defined as a minimax game with the following objective function:
\begin{equation}
    L_{adv}=E_{x}[log D_{src}(x)]+E_{x}[log(1-D_{src}(G(x)))],
\end{equation}
where the generator $G$ is trained to fool the discriminator $D$, while the discriminator $D$ tries to distinguish between generated samples $G(x,c)$ and real samples $x$.

In practice, training a GAN successfully is a notoriously difficult task that has given rise to many improvements. StarGAN \cite{choi2017stargan} has shown impressive results in image-to-image translation. Besides adversarial loss is used in training, attribute classification $L_{cls}$ and image reconstruction loss $L_{rec}$ are employed resulting in state-of-the-art attribute translation. The full objective is as following:
\begin{equation}
    L = L_{adv}+\lambda_{cls}L_{cls}+\lambda_{rec}L_{rec},
\end{equation}
We follow the same architecture in \cite{choi2017stargan} for facial attributes translation.

\section{Experiment}\label{sec:experiment}

\subsection{Datasets}\label{sec:datasets}

For beauty analysis, we deploy two rated datasets for experimental analysis. {\em Chen et al.} \cite{zhang2016new} built a beauty database with diversified and ethnic groups (we refer to Beauty 799). They collected $799$ female face images in total, $390$ celebrity face images including Miss Universe, Miss World, movie stars, and super models, and $409$ common face images.  They use a 3-point integer scale for rating: $3$ for unattractive, $2$ for common, and $1$ for attractive. Each image is rated by $25$ volunteers. Another dataset is the 10k US Adult Face Database \cite{bainbridge2013intrinsic}, which consists of $10168$ American adults, $2222$ faces are labeled on Amazon Mechanical Turk with $12$ respondents.
Different from rating on Beauty $799$ \cite{zhang2016new}, the 10k US Adult Face \cite{bainbridge2013intrinsic} use a 5-point integer attractiveness scale, $5$ represents the most attractive, $1$ is for most unattractive. Descriptions of these two datasets see in Table \ref{tab:dataset_desc}.

On attributes training stage, CelebA \cite{liu2015deep} is deployed for facial attribute estimation. There are $202,599$ images containing $10,177$ identities, each of which has $40$ attributes labels. Following their protocol \cite{liu2015deep}, we split the dataset into three folders: $160,000$ images of $8,000$ identities are used for training, and the images of another $20,000$ of $1,000$ identities are employed as validation. The remaining $20,000$ images of $1,000$ identities are used for testing. 

\subsection{Settings of Facial Attribute Training}\label{sec:settings} 

As Section \ref{sec:training} mentioned, we employ a pre-trained model of LightCNN \cite{wu2018light} as the feature extractor to perform facial attributes estimation.  Each face image is represented as a 256-D vector from fully connected layer of LightCNN and then fed into random forests along with the corresponding attribute labels for training.




Following the protocol of \cite{liu2015deep} , we are able to achieve $85\%$ accuracy averaging $40$ attributes estimation tested on CelebA \cite{liu2015deep}, which is comparable to state-of-the-art \cite{rudd2016moon} but more efficient due to no deep training. 

\begin{table*}
\begin{center}
\begin{tabular}{r|rllc}
\hline\hline
\textbf{Attribute} & \textbf{PCC} & \textbf{$p$-value (PCC)} & \textbf{$p$-value (Welch)} & \textbf{Correlation} \\
\hline
Arched Eyebrows     & $-0.109$ & $1.93 \times 10^{-3}$  & $1.46 \times 10^{-3}$   & Positive \\
Attractive          & $-0.208$ & $2.71 \times 10^{-9}$  & $7.17 \times 10^{-9}$   & Positive \\
Big Nose            & $0.054$  & $1.28 \times 10^{-1}$  & $4.82 \times 10^{-2}$   & Negative \\
Black Hair          & $0.062$  & $7.97 \times 10^{-2}$  & $9.00 \times 10^{-3}$   & Negative \\
Blond Hair          & $0.073$  & $3.80 \times 10^{-2}$  & $1.94 \times 10^{-2}$   & Negative \\
Bushy Eyebrows      & $0.034$  & $3.43 \times 10^{-1}$  & $1.61 \times 10^{-1}$   & -- \\
Heavy Makeup        & $-0.203$ & $6.77 \times 10^{-9}$  & $4.09 \times 10^{-9}$   & Positive \\
High Cheekbones     & $-0.107$ & $2.47 \times 10^{-3}$  & $1.06 \times 10^{-3}$   & Positive \\
Male                & $0.206$  & $4.28 \times 10^{-9}$  & $1.22 \times 10^{-9}$   & Negative \\
Mouth Slightly Open & $0.086$  & $1.55 \times 10^{-2}$  & $8.25 \times 10^{-3}$   & Negative  \\
No Beard            & $-0.040$ & $2.57 \times 10^{-1}$  & $7.21 \times 10^{-2}$   & -- \\
Smiling             & $0.005$  & $8.98 \times 10^{-1}$  & $4.50 \times 10^{-1}$   & -- \\
Wavy Hair           & $-0.062$ & $8.08 \times 10^{-2}$  & $5.83 \times 10^{-2}$   & -- \\
Wearing Earrings    & $-0.047$ & $1.90 \times 10^{-1}$  & $1.90 \times 10^{-1}$   & -- \\
Wearing Lipstick    & $-0.245$ & $2.43 \times 10^{-12}$ & $1.69 \times 10^{-12}$  & Positive \\
Young               & $-0.088$ & $1.25 \times 10^{-2}$  & $2.35 \times 10^{-3}$   & Positive \\
\hline\hline
\end{tabular}
\end{center}
\caption{Significant attributes tested in the \textbf{Beauty 799 dataset}. PCCs, $p$-values, and correlations are as discussed in Section \ref{sec:corr_analysis}.}
\label{tab:b799_results}
\end{table*}

\begin{figure*}[t]
\begin{center}
\includegraphics[width=0.9\textwidth]{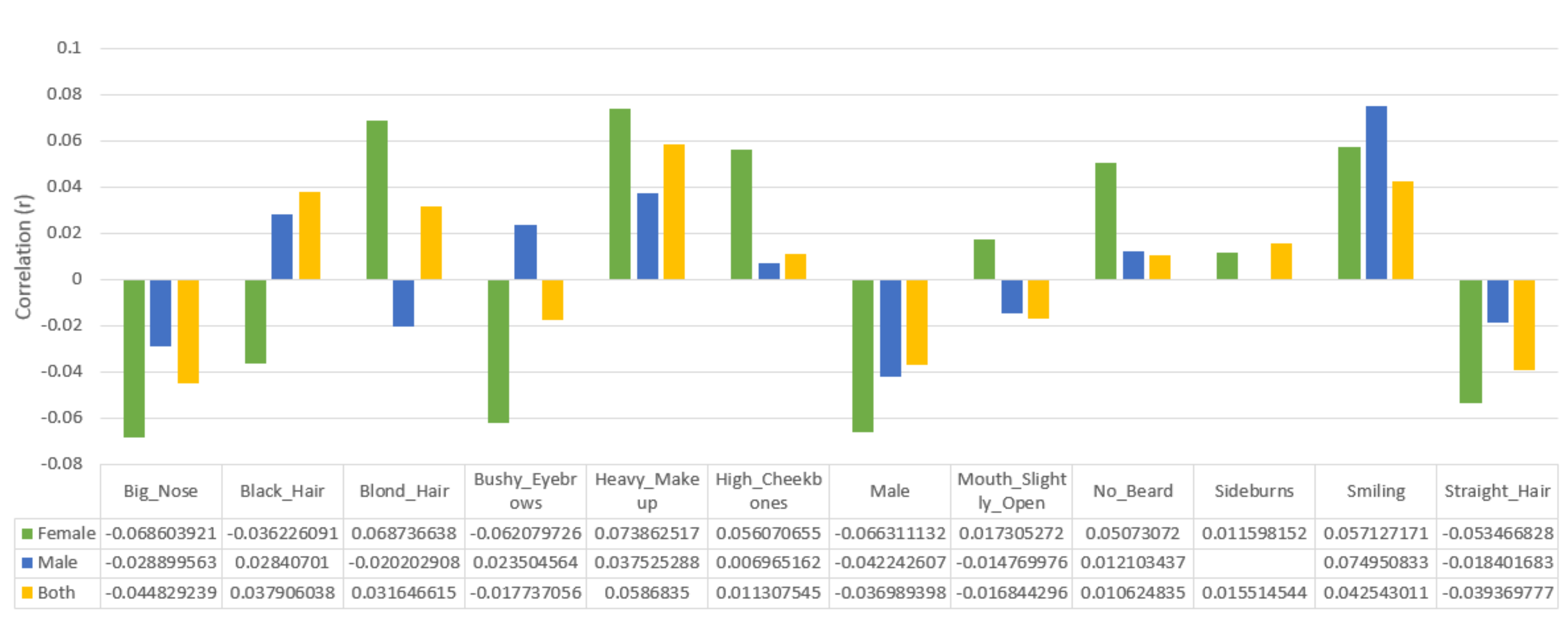}
\end{center}
\caption{Correlation analysis in 10K US database, including the subgroup of female, the subgroup of male, and the entire dataset.}
\label{fig:correlation}
\end{figure*}

\begin{table*}
\begin{center}
\begin{tabular}{r|rllc}
\hline\hline
\textbf{Attribute} & \textbf{PCC} & \textbf{$p$-value (PCC)} & \textbf{$p$-value (Welch)} & \textbf{Correlation} \\
\hline
Arched Eyebrows     & $0.055$  & $2.13 \times 10^{-3}$  & $1.60 \times 10^{-3}$   & Positive \\
Attractive          & $0.151$  & $8.27 \times 10^{-17}$ & $9.82 \times 10^{-17}$  & Positive \\
Big Nose            & $-0.047$ & $9.20 \times 10^{-3}$  & $2.77 \times 10^{-3}$   & Negative \\
Heavy Makeup        & $0.104$  & $1.10 \times 10^{-8}$  & $1.17 \times 10^{-8}$   & Positive \\
High Cheekbones     & $0.043$  & $1.80 \times 10^{-2}$  & $9.56 \times 10^{-3}$   & Positive \\
Male                & $-0.081$ & $8.18 \times 10^{-6}$  & $6.03 \times 10^{-6}$   & Negative \\
Mouth Slightly Open & $-0.035$ & $5.16 \times 10^{-2}$  & $2.45 \times 10^{-2}$   & Negative \\
Wearing Lipstick    & $0.122$  & $1.60 \times 10^{-11}$ & $3.74 \times 10^{-11}$  & Positive \\
Young               & $0.042$  & $1.97 \times 10^{-2}$  & $3.63 \times 10^{-3}$   & Positive \\
\hline\hline
\end{tabular}
\end{center}
\caption{Significant attributes tested in the \textbf{combined dataset}. PCCs, $p$-values, and correlations are as discussed in Section \ref{sec:corr_analysis}.}
\label{tab:combined_results}
\end{table*}

\subsection{Attribute Selections for Attractiveness}\label{sec:correlation_exp}

After obtaining high-level facial representations by above attributes training, we investigate correlations between these attributes and beauty scores from the labelled datasets. For each entry in the datasets, the attribute is binary (either $0$ or $1$) and the beauty score is decimal ranging from $1$ to $5$ (10K US dataset) or from $1$ to $3$ (Beauty 799 dataset).

To better understand gender differences, we split the 10K US dataset into three folders: female, male, and both. We also normalize the beauty scores using standard score \cite{zill2011advanced} and merge two datasets into a larger one to resolve missing data issues of some entries. Five subsets (i.e., Beauty 799, 10K US, 10K US for female, 10K US for male, and the combination) are then given. As discussed in Section \ref{sec:corr_analysis}, we calculate Pearson's correlation coefficients and perform significance tests on aforementioned subsets respectively. Correlation coefficients, selection decisions, and corresponding $p$-values of both coefficient significance tests and single-tailed two-sample Welch's $t$-tests are reported in Table \ref{tab:combined_results},
where {\em Positive}, {\em Negative}, and {\em $-$} indicate positive linear relationship, negative linear relationship, and not significance respectively.

\begin{figure}[t]
\begin{center}
\includegraphics[width=0.5\textwidth]{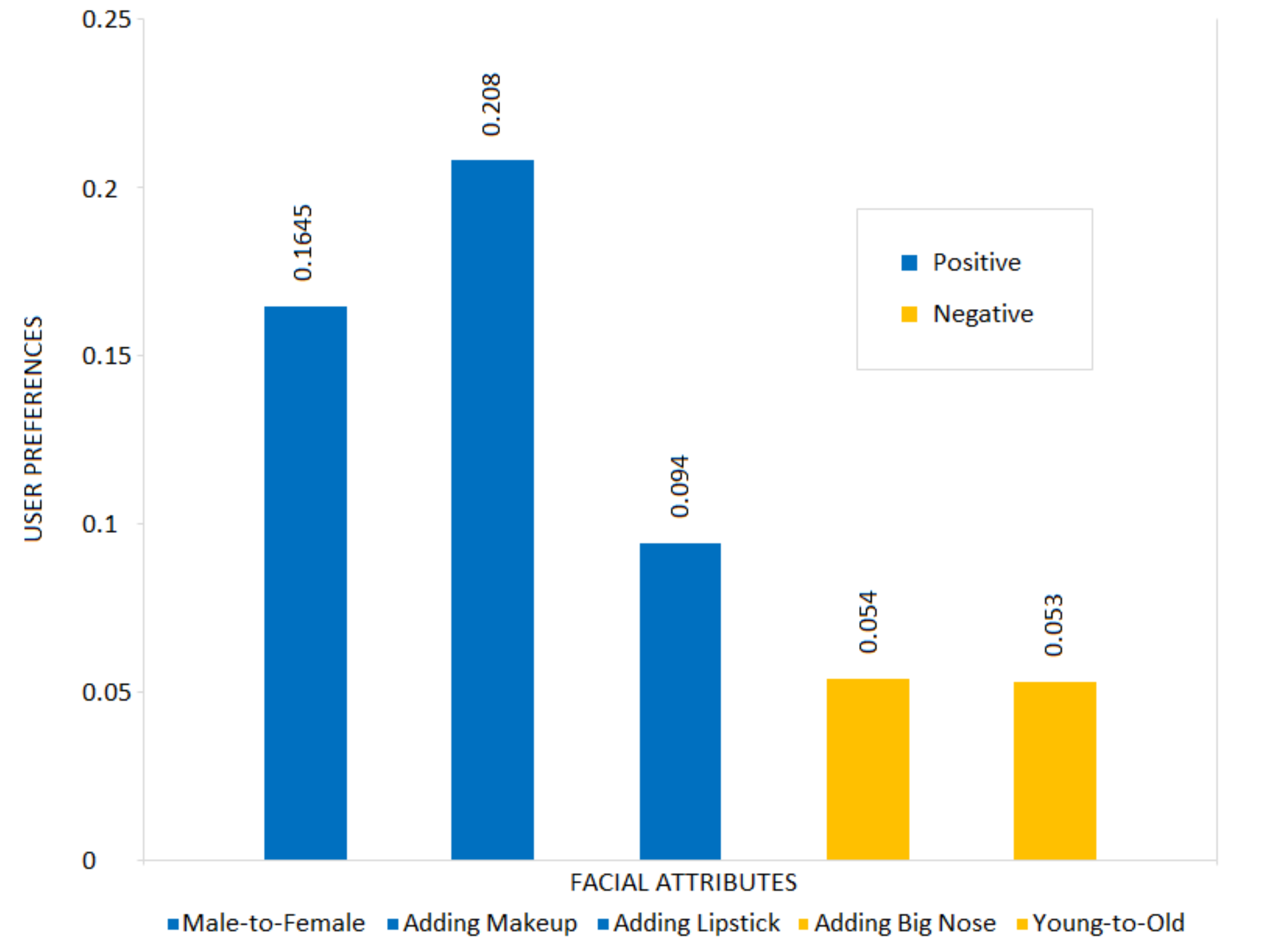}
\end{center}
\caption{User study result verifies our hypotheses and correlation analyses. The percentage of the user preferred choices over all images(vertical axis) ranks five translation options in the following order: adding Heavy Makeup, Male-to-Female, adding Lipstick, adding Big Nose and Young-to-Old.}
\label{fig:user_study}
\end{figure}

\subsection{Attributes Translation for Beauty Evaluation}\label{sec:change}

After mining semantics of beauty by correlation and significance testing, additional experiments are made by changing facial attribute to evaluate beauty differences.

StarGAN \cite{choi2017stargan} has shown impressive results on image-to-image translation. In this experiment, we employ StarGAN as the architecture for facial attributes translation. Similar protocol to StarGAN, CelebA database is deployed for attributes translation. To evaluate the subjective differences in beauty, we conduct a perceptual study on Amazon Mechanical Turk (AMT). Five translation options, associated with three positive and two negative facial attributes, are assessed: Male-to-Female, adding Heavy Makeup, adding Lipstick, adding Big Nose and Young-to-Old. The original and five translated images from 50 CelebA identities are used. Thus, there is a total of 250 pairs of images in the study. Each participant chooses between 50 pairs of images and selects the face they find better looking in each pair. The two images of a pair are selected from the original image of an identity and one of its five translated images, presented side-by-side. Each pair of images are assessed by 40 participants. We finally obtained 10000 valid assessments.

The participants' preferences are analyzed using logistic regression, a statistical model commonly used for binary outcome variables. The subjective results of the five facial attributes are presented in Figure \ref{fig:user_study}. Planned comparisons reveal that the participants preferred the five translation options in the following order: adding Heavy Makeup was the most preferable, Male-to-Female the second, adding Lipstick the third; adding Big Nose and Young-to-Old were both the least preferable (i.e. no significant difference between Big Nose and Young-to-Old). The user study result aligns well with our hypotheses and correlation analyses discussed in Section \ref{sec:corr_analysis}.

\section{Analysis}\label{sec:analysis}

\subsection{Beauty Semantics on Beauty 799 Dataset}\label{sec:b799}

Beauty 799 dataset \cite{zhang2016new} only consists of images of females and scores are $1$, $2$, or $3$, indicating {\em very attractive}, {\em common}, or {\em unattractive} respectively. In Section \ref{sec:corr_analysis}, we have discussed ways to determinate correlations. Take Arched Eyebrows as an example, its $r$ equals $-0.109$ which indicates that Arched Eyebrows has a negative correlation with the beauty score (Y). Since Arched Eyebrows only can be chosen 0 or 1, specifically, it indicates when people have the attribute of Arched Eyebrows (1), the beauty score (Y) is going down, but small beauty score (Y) represents more attractive (refer to original rating). Therefore, the attributes with negative $r$ have a positive impact on beauty.
As a result, as shown in Table \ref{tab:b799_results}, we are able to generate all the correlations between facial attributes and beauty degree on Beauty 799 \cite{zhang2016new}.

From Beauty 799 dataset, first, we can conclude that people who have such attributes, like, Arched Eyebrows, Makeup, High Cheekbones, Wavy Hair, Wearing Earrings, Wearing Lipstick, Young, are more attractive. On the other hand, it is recognized as less attractive for these attributes, such as Big Nose, Black Hair, Blond Hair, Male, Mouth Slightly Open.

\subsection{Beauty Semantics on 10k US Dataset}\label{sec:us10k}

Different from Beauty 799, the 10k US Adult Face Database \cite{bainbridge2013intrinsic} contains more images and consists of both males and females but only Americans. The scales of beauty score in 10k US Adult Face Database \cite{bainbridge2013intrinsic} are five levels, and 1 indicates the least attractive, 5 indicates the most attractive.The correlation between beauty score and attribute feature is computed by Pearson Correlation as shown in Figure \ref{fig:correlation}, and positive correlation suggests people with these attributes have a positive impact on beauty in this dataset.

As previously mentioned, we divide three parts for analyzing the beauty semantics in the 10k US dataset \cite{bainbridge2013intrinsic}. When considering the whole dataset including both female and male (see in Figure \ref{fig:correlation}), the attributes with Black Hair, Heavy Makeup, High Cheekbone, No Beard, Smiling and Wearing Lipstick are positive to a person's beauty. On the other hand, the attributes including Big Nose, Blond Hair, Bushy Eyebrows, Male, Mouth Slightly Open, Straight Hair as well as Young have negative impacts on beauty. That is the general beauty semantics conclusion on the 10k US.

More specifically, when experimenting the beauty semantics only using female face images, Blond Hair and Sideburns are considered as the positive effect on beauty. On the other hand, apart from the general negative attributes(Big Nose, Blond Hair, Bushy Eyebrows, Male, Mouth Slightly Open, Straight Hair) generated from the entire dataset of the US 10k, the attributes with Black Hair and Bushy Eyebrows for female are considered as negative attributes to beauty. Meanwhile, when studying the beauty of male, we find out all those attributes which would enhance beauty still play positive roles in beauty except Blond Hair, instead, Blond Hair is considered as an unattractive attribute.

\subsection{Feminine Features for Beauty}\label{sec:female}

Not only are we able to conclude the objective beauty semantics using data statistics, but there is another interesting finding that feminine features are recognized as more attractive compared to masculine features. From psychological perspective, there are considerable evidences that feminine features increase the attractiveness of male and female faces across different cultures \cite{little2002role,perrett1998effects,rhodes2000sex,little2001self,little2002partnership}. Applied to our experiment, attributes like Heavy Makeup and Wearing Lipsticks are generally considered as feminine feature. Therefore, it is a consistent interpretation that these attributes have a positive effect on attractiveness both from our statistical results and psychology. Besides, there is a gender attribute named Male of which the prediction is convincing tested in CelebA from our model (96.5\% accuracy). However, we found an interesting result that some females are estimated as males from model outcomes in Beauty 799 database, which indicates those females carrying some masculine features (Male tendency) are recognized as less attractive. Furthermore, this Male bias attribute decreases the attractiveness from the correlation analysis. That is a contrary evidence that turns out feminine features increase the attractiveness based on our finding.

\subsection{Inconsistent and Identical Semantics}\label{sec:same}

As aforementioned, there are some intrinsic differences between these two databases \cite{zhang2016new,bainbridge2013intrinsic}. As a result, beauty semantics are not exactly the same. Some interesting findings are illustrated: the US adults have a preference on Black Hair and Blond Hair, which turns out an opposite conclusion to the results from Beauty 799. This phenomenon might be affected by environment, different culture may have slight preferences for hair color and shape. Moreover, apart from the inconsistency crossing database, in 10k US database, experiments illustrate that Black Hair and Bushy Eyebrows are attractive attributes referring to the male. However, it is an absolute reverse when it comes to female results, both Black Hair and Bushy Eyebrows have negative effects on beauty understanding. Another inconsistent attribute is Blond Hair between females and males, for females it is recognized as a positive attribute on beauty, but it is negative for males.

Even some inconsistencies occur in \cite{zhang2016new,bainbridge2013intrinsic}, there still exists some identical semantics for both positive and negative on attractiveness in \cite{zhang2016new,bainbridge2013intrinsic}. The attributes that identically play a positive or negative role in beauty from these two relatively large datasets are summarized in Table \ref{tab:combined_results}. For example, these attributes: Heavy Makeup, High Cheekbones,  Wearing Lipstick would increase attractiveness (Beauty). Instead, the attributes with Big Nose, Male bias (refer to the female), and Mouth Slightly Open have a negative impact on attractiveness.

\section{Conclusion}\label{sec:conclusion}

We propose a method for understanding beauty via deep facial features. Our contribution is discovering facial attributes with significant positive or negative impact to attractiveness verified by statistical tests. Our study not only provides quantitative evidence for psychological beauty studies, but more significantly, reveals the high-level features for beauty understanding which are critical for beauty enhancements. We further manipulate several facial attributes with a GAN based approach, and validate our findings with a large-scale user survey. Our study is the first attempt to understand beauty, highly perceptual to human, via a deep learning perspective. This opens up many opportunities for interdisciplinary research as well as applications.




{\small
\bibliographystyle{IEEEtran}
\bibliography{db}
}

\end{document}